\documentclass[lettersize,journal]{IEEEtran}
\usepackage{amsmath,amsfonts}
\usepackage{algorithmic}
\usepackage{algorithm}
\usepackage{array}
\usepackage[caption=false,font=normalsize,labelfont=sf,textfont=sf]{subfig}
\usepackage{textcomp}
\usepackage{stfloats}
\usepackage{url}
\usepackage{verbatim}
\usepackage{graphicx}
\usepackage{cite}
\usepackage{multirow}
\usepackage{color}
\usepackage{algorithm, algorithmic}

\begin{document}

\title{RBA-GCN: Relational Bilevel Aggregation Graph Convolutional Network for Emotion Recognition }

\author{
  \IEEEauthorblockN{Lin Yuan, Guoheng Huang$^{*}$, Fenghuan Li, Xiaochen Yuan$^{*}$,  Chi-Man Pun, Guo Zhong$^{*}$}

}

\markboth{}%
{Shell \MakeLowercase{\textit{et al.}}: A Sample Article Using IEEEtran.cls for IEEE Journals}



\maketitle

\begin{abstract}
Emotion recognition in conversation (ERC) has received increasing attention from researchers due to its wide range of applications. As conversation has a natural graph structure, numerous approaches used to model ERC based on graph convolutional networks (GCNs) have yielded significant results. However, the aggregation approach of traditional GCNs suffers from the node information redundancy problem, leading to node discriminant information loss. Additionally, single-layer GCNs lack the capacity to capture long-range contextual information from the graph. 
Furthermore, the majority of approaches are based on textual modality or stitching together different modalities, resulting in a weak ability to capture interactions between modalities. To address these problems, we present the relational bilevel aggregation graph convolutional network (RBA-GCN), which consists of three modules: the graph generation module (GGM), similarity-based cluster building module (SCBM) and bilevel aggregation module (BiAM). First, GGM constructs a novel graph to reduce the redundancy of target node information. Then, SCBM calculates the node similarity in the target node and its structural neighborhood, where noisy information with low similarity is filtered out to preserve the discriminant information of the node. Meanwhile, BiAM is a novel aggregation method that can preserve the information of nodes during the aggregation process. This module can construct the interaction between different modalities and capture long-range contextual information based on similarity clusters. On both the IEMOCAP and MELD datasets, the weighted average F1 score of RBA-GCN has a 2.17$\sim$5.21\% improvement over that of the most advanced method. Our code is available at https://github.com/luftmenscher/RBA-GCN and our article  "RBA-GCN: Relational Bilevel Aggregation Graph Convolutional Network for Emotion Recognition"  was published in IEEE/ACM Transactions on Audio, Speech, and Language Processing, vol.31, pp.2325-2337, 2023, doi: 10.1109/TASLP.2023.3284509.

\end{abstract}

\begin{IEEEkeywords}
Emotion recognition, multimodal fusion, context modeling, similarity cluster.
\end{IEEEkeywords}

\section{Introduction}
\IEEEPARstart{T}{he} purpose of emotion recognition in conversation (ERC) is to assign each sentence in a conversation to a specific emotion category. ERC is becoming an important research topic due to its broad applications in various scenarios, such as chatbots and mental health services \cite{31,32}. Cambria et al. \cite{56} consider understanding emotions to be an important aspect of personal development and growth; as such, it is key for the emulation of human intelligence.

\begin{figure}[t]
    \centering
    \includegraphics[width=9cm]{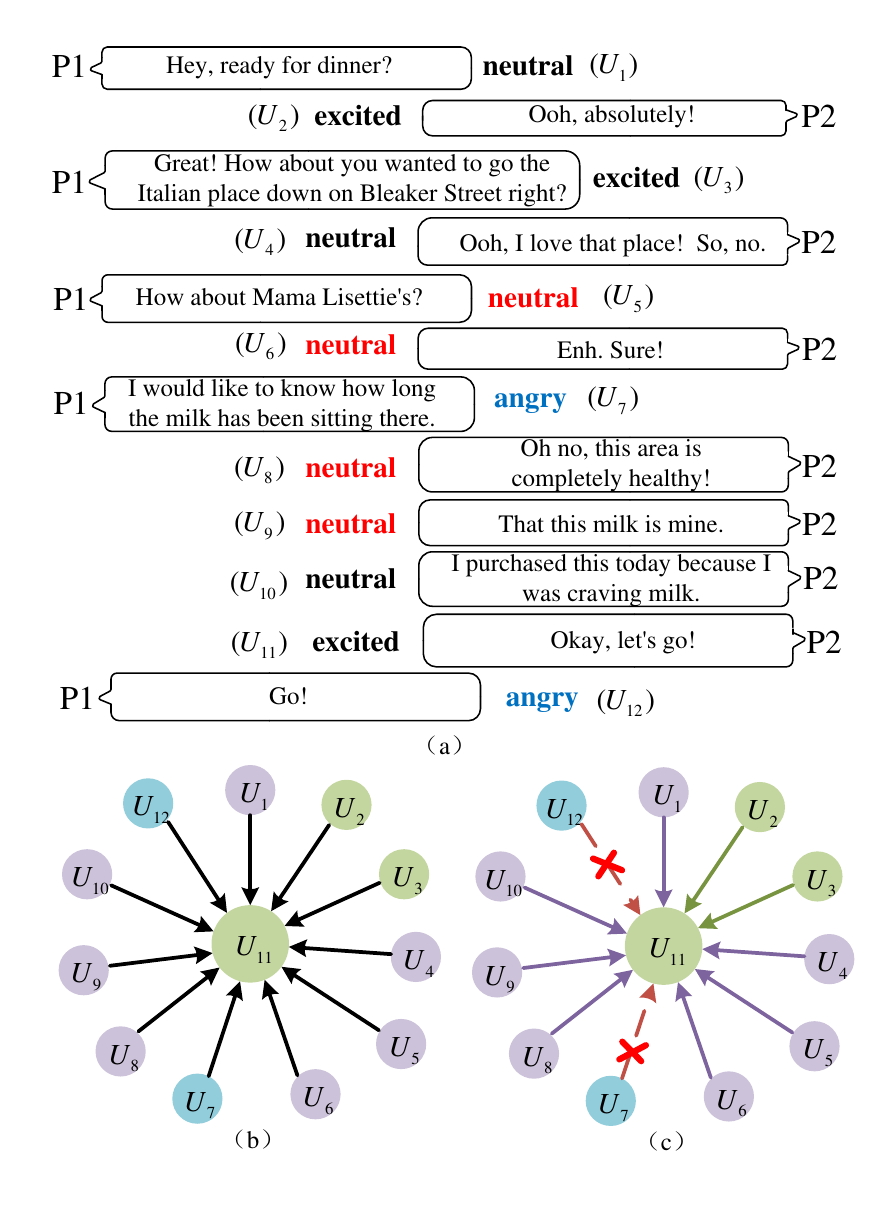}
    \caption{ (a) An example of a conversation with different aggregation methods. The emotion will be predicted for the sentence with blue label.  Traditional graph convolutional network would aggregates the blue-labeled sentence with the red-labeled sentences (neighboring nodes). The two blue-labeled sentences ($U_7$ and $U_{12}$) are aggregated together by our method. (b) (c): Take $U_{11}$ as the target node as an example. Different colors of nodes represent different labels. (b) is the traditional graph convolution method, which aggregates node information in the graph without difference. (c) is our propose method, which performs bilevel aggregation based on the clusters. Different colors of edges represent the other cluster. The dashed arrows indicate the filtered nodes. }
    \label{fig1}
\end{figure}

The ERC task differs from traditional emotion recognition of individual isolated utterances in that it requires a combination of conversational intent, topic and context \cite{43}. Previous models are mainly 
tested 
by means of contextual information, e.g., bias compensation-long short-term memory (BC-LSTM) \cite{49}, conversational memory network (CMN) \cite{1}, and dialogue recurrent neural network (DialogueRNN) \cite{3}. However, these models cannot effectively capture long-range contextual information in a multiperson conversation scenario. To address the shortcomings of the above models, some ERC models based on graph convolutional networks (GCNs), such as DialogueGCN \cite{4} and multimodal fusion via deep graph convolution network (MMGCN) \cite{5}, have been proposed. The DialogueGCN model captures the dependencies between speakers by forming utterances in a conversation into a fully connected graph. Different from DialogueGCN, which utilizes only textual information, MMGCN further leverages multimodal information for emotion recognition. Similarly, it uses all the different modalities of utterances in a conversation as nodes to form a fully connected graph and applies multilayer GCNs to capture long-range contextual information.

Although previously developed ERC methods have achieved great progress, they mainly exploit GCNs based on message passing neural networks (MPNNs) \cite{6,11}. 
 Consequently, such models possess several shortcomings. First, single-layer GCNs aggregate only neighboring nodes. In a conversation, utterance nodes that are far from each other may also have high structural similarity. However, due to the influence of graph generation methods, a single-layer GCN may be unable to capture such utterance node information. To solve this problem, multilayer GCNs are often used to capture long-range contextual information. However, GCNs simply sum the “messages” from all neighborhoods. After aggregating the neighboring information via multilayer GCNs, the information possessed by similar nodes at distant locations may be disturbed by a large amount of irrelevant noisy information acquired from the nodes that are proximal to the prediction target. This leads to a situation where long-range contextual information cannot be efficiently extracted.
 Velickovic et al. \cite{54}. and Ishiwatari et al. \cite{25} adopted an attention mechanism to reduce the interference of irrelevant noise information by assigning corresponding weights to adjacent nodes. In contrast, we take a different approach. We utilize the cosine similarity function to calculate the similarity between nodes, filter nodes with low similarity, and then map them to corresponding clusters according to their similarity levels. With this approach, we can effectively eliminate redundant information and preserve the discriminant information of the node.
As shown in Figure \ref{fig1} \footnote{“Friends” Season 5 ep7: http://www.livesinabox.com/friends/scripts.shtml}, we first consider long-range contextual information. Although $U_2$ and $U_{11}$ are far away, they both express excitement because they are related to the topic of eating, which can illustrate the importance of long-range contextual information for ERC. 
The traditional aggregation methods indiscriminately aggregate the target node $U_7$ and its neighboring nodes $U_6$ and $U_8$ . In contrast, our method first filters the redundant information. Then, the information within each cluster is aggregated. Finally, the information between clusters is aggregated, thus avoiding the disturbance caused by the noise of the proximal nodes and better preserving the discriminant information of the target node. Here, we define the target node as the node in the graph that currently needs to be predicted.

In summary, a relational bilevel aggregation graph convolutional network (RBA-GCN) is presented in this paper, which can capture long-range contextual information in a single-layer architecture and improve the ability to capture interactions between different modalities. Different from DialogueGCN and MMGCN, we leverage the disconnected neighborhood to handle long-range contextual information and the connected neighborhood to handle multimodal interactions. First, we model the contextual information via bidirectional long short-term memory (Bi-LSTM) with the extracted features of different modalities. Based on this, we propose to connect nodes of the same modality in the same conversation in order of conversation and connect different modalities in the same utterance. 
We compute the similarity between the target node and the nodes in its structural neighborhood by corner similarity and map these nodes to different clusters.
In particular, we remove the nodes with low similarity in the relation definition to effectively filter out the interference of noisy information. To allow RBA-GCN to be applied to input data in different orders, making the model more robust and general, we introduce the design consideration of permutation invariance. To ensure the permutation invariance of the graph-structure data, we utilize the bilevel aggregation module (BiAM) to renew the feature representation of the node, thereby generating the final classification features of the target node. Finally, we pass the final classification features of the target node through an emotion classifier to facilitate emotion prediction.

The contributions of this paper can be summarized as follows:

\begin{itemize}
    \item A novel ERC framework (RBA-GCN) is proposed to comprehensively consider the relevance between nodes on the basis of graphs. The proposed RBA-GCN can capture long-range context information and interactions between modalities under a single-layer architecture.
    \item To reduce the redundancy of the target node information, we present a novel graph generation module (GGM). Based on the GGM, we propose the similarity-based cluster building module (SCBM), which considers the correlation between nodes, to enhance the interclass relationship based on the similarity metric.
    \item We present a novel graph convolution aggregation method, BiAM, to aggregate the feature representations of distant nodes through a cluster neighborhood and perform multimodal feature fusion. The proposed BiAM can preserve the discriminant information of nodes during the aggregation process.
    \item To verify the performance of our approach, experiments on both the IEMOCAP and MELD datasets are conducted. On both datasets, the weighted average F1 score of our approach is improved by 2.17$\sim$5.21\% over that of the state-of-the-art method.
\end{itemize}

\begin{figure*}
    \centering
    \includegraphics[width=18cm,page={1}]{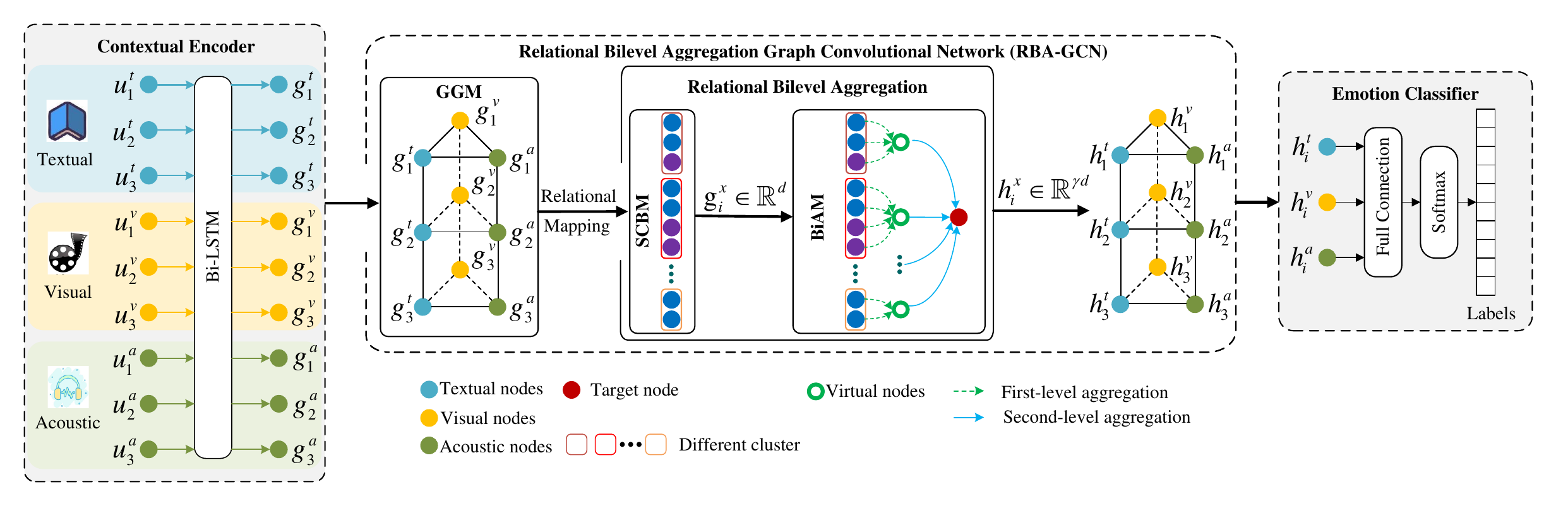}
    \\ \hspace*{\fill} \\
    \caption{ The overall framework. First, we encode contextual information for each modality feature of the utterance, using Bi-LSTM to obtain the contextual embedding of each node. Then, we apply the  RBA-GCN to filter out noisy information and reduce information redundancy, while effectively capturing the interactions between modalities and long-range contextual information. Finally, classifiers are applied to implement emotion prediction.}
    \label{fig2}
\end{figure*}

\section{Related Work}
In this section, we briefly introduce  recent deep learning-based methods for ERC tasks \cite{52}. The specific methods are described as follows:

\textbf{Contextual Modeling Emotion Recognition:} 
Several advances have been made in ERC research, as the number of open-source datasets available for ERC has increased \cite{18,20}. First, Hazarika et al. \cite{1} presented a CMN that utilized the different memories of each speaker to model the specific context of the speaker. Second, Hazarika et al. \cite{2} presented an interactive conversational memory network (ICON) that accounted for the influence of interpersonal relationships in a conversation and modeled the affective influence between self and speaker hierarchically as a global memory. Then, Majumder et al. \cite{3} presented the DialogueRNN, which utilized three GRU modules interacting with each other to model conversational information. In recent years, due to the outstanding ability of GCNs to process contextual information, GCNs have been used extensively in emotion recognition. For example, DialogueGCN \cite{4} constructs a fully connected graph by referring to each utterance in the conversation, while the edges between two nodes form speaker dependencies. This is the first GCN-based model for emotion recognition, and good results have been obtained. Tu et al. \cite{58} presented a context and emotion-aware framework, termed Sentic GAT, which tends to select common sense knowledge consistent with the context semantics and emotion of the target utterance. This approach has also achieved good results. Finally, since the DialogueGCN considers only the textual modality, the MMGCN \cite{5} builds upon it by exploiting information from multiple modalities and encoding the information of the speaker. This method uses multilayer GCNs to capture long-range contextual information and realizes the best performance. The above GCNs applied to ERC are all part of MPNNs. To capture long-range contextual information, the multilayer GCN strategy is typically used. 
However, when applying multilayer GCNs, more computer resources are consumed, and the updated node information after aggregation contains a considerable amount of irrelevant information. Thus, the discriminant information of the target node is lost, and the ability to capture the context information remotely is diminished. To resolve the problems in the ERC mission, we present an RBA-GCN, which is inspired by the GEOM-GCN \cite{9}. GEOM-GCN maps nodes into a continuous latent space, followed by the construction of a structural neighborhood for aggregation using the geometric relationships defined in the latent space.

\textbf{Multimodal Emotion Recognition:}
Based on textual modality development and the increasing number of multimodal emotion recognition datasets \cite{7,8}, more researchers have been focusing on the exploitation of multimodal information. Hazarika et al. \cite{21,22} simply concatenated the features of the three modalities in series for multimodal fusion with no established intermodal interactions. Chen et al. \cite{23} performed word multimodal fusion for emotion recognition in solitary utterances. Zadeh et al. \cite{24} proposed an MFN to fuse multiview information, which can satisfactorily coordinate features of different modalities. However, the feature fusion technique of these methods is the simple splicing of features \cite{40,41}. Lian et al. \cite{47} proposed CTNet using a transformer-based structure to model fusion between multimodal features. Chen et al. \cite{45} proposed a novel time and semantic interaction network (TSIN) to conduct emotional parsing and emotion refinement by performing fine-grained temporal alignment and cross-modal semantic interaction. Although these methods achieve some improvement in performance, the problem of data sparsity can easily occur with high-dimensional features \cite{39}. Recently, Zhang et al. \cite{59} proposed a novel multimodal emotion recognition model for conversational videos based on reinforcement learning and domain knowledge (ERLDK); this model introduces reinforcement learning algorithms for real-time ERC with the occurrence of conversations. Yang et al. \cite{60} proposed a multimodal framework named two-phase multitask emotion analysis (TPMSA). This method applies a two-stage training strategy to leverage pretrained models and a novel multitask learning strategy to investigate classification capabilities. In contrast to existing studies, our proposed graph method can preserve multimodal information and effectively capture the interactions between modalities.

\section{Proposed Method}

Our approach is described throughout this section. The framework of our proposed model, which is displayed in Figure \ref{fig2}, is composed of a contextual encoder, an RBA-GCN, and an emotion classifier. In the contextual encoder part, the extracted features are passed into the Bi-LSTM layer to generate contextual information of the utterance. Then, the proposed RBA-GCN is applied to capture both long-range contextual information and multimodal information. The information of nodes is preserved during the aggregation. In the emotion classifier part, the node features updated by RBA-GCN are used as features for the final classification.

\subsection{Problem Definition}\label{subsec3}

First, a series of utterances $ \left \{ u_{1},u_{2},\ldots,u_{N} \right \} $ composes a conversation, where $ N $ represents the number of utterances in a conversation. The 
objective 
of ERC is to identify emotional labels (“happy”, “excited”, “sad”, “frustrated”, “neutral”, “angry”) for each utterance. Each utterance contains three modalities of data, namely, textual ($ t $), visual ($ v $), and acoustic ($ a $), which are represented as follows:

\begin{equation}
u_{i}=\left \{ \boldsymbol{u}_{i}^t,\boldsymbol{u}_{i}^v,\boldsymbol{u}_{i}^a \right \} 
\end{equation}
\noindent
where $ \boldsymbol{u}_{i}^t $, $ \boldsymbol{u}_{i}^v $, and $ \boldsymbol{u}_{i}^a $ represent the original feature representations of the textual, visual, and acoustic modalities of utterance $ u_{i} $, respectively.

\subsection{Contextual Encoder}\label{subsec3}

Context refers mainly to factors such as time, occasion, and place in which language activities occur. Contextual information is essential for ERC, especially during some short utterances, which are very important for predicting emotional labels. Therefore, we encode contextual information for each modality feature of the utterance. We input the per-modality features of an utterance into a Bi-LSTM network to encode orderly contextual information of each modality. The contextual information feature encoding is implemented as follows:

\begin{equation}
\boldsymbol{g}_{i}^{x}=\left[\overrightarrow{\operatorname{LSTM}}\left(\boldsymbol{u}_{i}^{x}, \overrightarrow{\boldsymbol{g}_{i-1}^{x}}\right), \overleftarrow{\operatorname{LSTM}}\left(\boldsymbol{u}_{i}^{x}, \overleftarrow{\boldsymbol{g}_{i+1}^{x}}\right)\right]
\end{equation}
\noindent
where $\boldsymbol{u}_{i}^{x}$ represents a context-independent arbitrary modality raw feature representation for utterance $i$ and $x \in \left \{ t,v,a \right \} $ represents an arbitrary modality of an utterance. $\overrightarrow{\boldsymbol{g}_{i-1}^x}$ is the hidden vector obtained before processing the current sentence, and $\overleftarrow{\boldsymbol{g}_{i+1}^x} $ is obtained after processing the current sentence.

After the original features pass through the Bi-LSTM network, the context encoder outputs context-aware feature encodings $\boldsymbol{g}_{i}^{t}$, $\boldsymbol{g}_{i}^{v}$, and $\boldsymbol{g}_{i}^{a}$ accordingly.

\subsection{Relational Bilevel Aggregation GCN (RBA-GCN)} \label{subsec3}

Our proposed RBA-GCN can filter out noisy information and reduce the redundancy of target node information. Long-range contextual information and interactions between modalities can be effectively captured. RBA-GCN consists of three modules: GGM, SCBM, and BiAM.

\subsubsection{Graph Generation Module (GGM)}\label{subsubsec3}

Previous graph convolution models for emotion recognition typically construct all nodes in a conversation as a fully connected graph. However, this method has the following drawbacks: First, in this approach, the graph network is very large, which makes the training of the model difficult. To address this problem, sliding windows are used by the models of the graph generation approach to aggregate and update target node, but the ability to capture long-range contextual information is lacking. Second, GCNs simply sum all the “messages” connected to the target node, whereas construction using fully connected graphs leads to redundant node information. Thus, we do not know which nodes contribute to the final aggregation. To address these issues, we adopt an effective graph generation method. The specific implementation details are as follows:

We construct each conversation containing $ N $ utterances as an undirected graph $\mathcal{G}=({V}, {E}) $, where $ {V}\left (\lvert{V} \rvert = 3N \right ) $ represents the nodes of three modalities in each utterance. $ {E}$ represents the edges between every two relation nodes. The graph is constructed as follows:

\noindent
\textbf{Nodes:} We represent each modality of each utterance in the conversation as a node of a graph, and the nodes of the three modalities of each utterance are represented as $ n_{i}^t, n_{i}^v $ and $ n_{i}^a $. The nodes are initialized with the outputs from the contextual encoders: $ \boldsymbol{g}_{i}^t,\boldsymbol{g}_{i}^v $ and $ \boldsymbol{g}_{i}^a $. Therefore, for a conversation with $N$ utterances, the graph has 3$N$ nodes.

\noindent
\textbf{Edges:} To exploit multimodal information more effectively and capture long-range contextual information, we connect nodes of the same modality in the conversation sequentially according to the conversation order. Nodes of several modalities of the same utterance are connected in the same conversation. For example, in the graph, we connect $n_{i}^t$, $n_{i}^v$ and $n_{i}^a$ to each other.

\subsubsection{Similarity-Based Cluster Building Module (SCBM)}\label{subsubsec3}

We first calculate the node similarity in the target node and its structure neighborhood. We consider nodes with low similarity to the target node to have opposite or different labels from the target node, and such nodes are filtered. Nodes with high similarity to the target node are considered to have similar features or the same label as the target node. We map these nodes to different clusters based on the similarity between the nodes.

\begin{figure}[t]
    \centering
    \includegraphics[width=9cm]{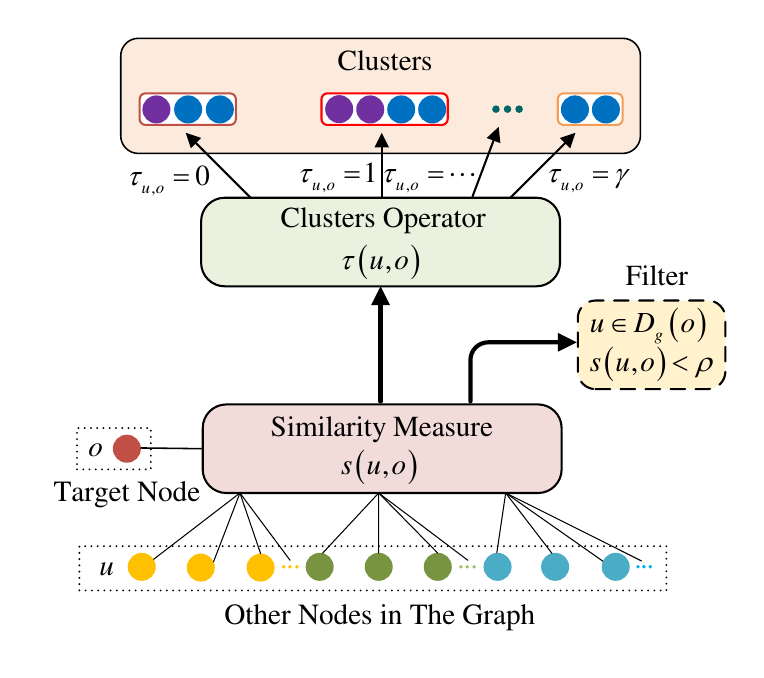}
    \\ \hspace*{\fill} \\
    \caption{\centering The construction process of similarity clusters.}
    \label{fig4}
\end{figure}

In this paper, we leverage the disconnected neighborhood to 
handle
long-range contextual information and the leverage connected neighborhood to handle multimodal interactions. First, we construct the structural neighborhood $N\left (o\right )$ on the basis of the GGM. Second, for the relationship between two nodes, we assume that the higher the similarity between them, the more similar the information between them and the higher the level of the relationship. Nodes in the same cluster have a certain similarity, and we believe that the aggregation operations of nodes in the same cluster can have a certain feature enhancement effect. Thus, we define the structural neighborhood $N\left (o\right )$ as follows:

\begin{equation}
N\left (o\right ) =\left ( \left \{ C_{g}\left (o\right ),D_{g}\left (o\right )  \right \}\right )
\end{equation}

\noindent
where $C_{g}\left (o\right )$ is the connected neighborhood in the graph and $D_{g}\left (o\right )$ is the disconnected neighborhood in the graph.

The connected neighborhood $C_{g}\left (o\right )$ in the graph is defined below:

\begin{equation}
C_{g}\left (o\right )= \left \{u\lvert u\in {V},\left ( u,o\right )\in {E}\right \}
\end{equation}

The disconnected neighborhood $D_{g}\left (o\right )$ in the graph is defined below:

\begin{equation}
D_{g}\left (o\right )=\left \{u\lvert u\in {V},\left ( u,o\right )\not \in {E}\right \}
\end{equation}

\noindent
where $u$ and the target node $o$ belong to the same modality.

The similarity metric function $s\left ( u,o \right )$ is defined below: 

\begin{equation}
s\left ( u,o \right )=\left(1-\frac{\arccos \left(\operatorname{sim}\left(\boldsymbol{f}_u, \boldsymbol{f}_o\right)\right)}{\pi}\right)\left ( u \in N\left (o\right ) \right )  
\end{equation}

\noindent
where $\operatorname{sim}\left(\cdot,\cdot \right)$ is the cosine similarity function. $\boldsymbol{f}_u$ and $\boldsymbol{f}_o$ represent the features of nodes $u$ and $o$ on the graph, respectively.

We define the cluster operator $\tau$ through similarity mapping and specifically define the clusters as follows:

\begin{equation}
\begin{split}
\tau\left( u,o\right) &= \left \lfloor \gamma\times s\left (  u,o \right ) \right \rfloor \text { if } u \in C_{g}\left (o\right)\\
& or \;  \left (   u \in D_{g}\left (o\right ) \cap  s\left (  u,o \right ) \ge\rho\right )
\end{split}
\end{equation}

\begin{equation}
Clusters=\left \{ Clusters^{\left ( r \right )} \mid \tau\left ( u,o \right )=r  \right \} 
\end{equation}

\noindent
where $\gamma$ and $\rho$ are hyperparameters. $\rho$ is the threshold value for filtering noisy information from the clusters. $\lfloor \cdot \rfloor$ is the rounding down operation. When the similarity is less than $\rho$ and $u$ is in the disconnected neighborhood, we filter out this node to reduce information redundancy. We set $\gamma$ to an integer so that we can obtain $\gamma$+1 clusters. $Clusters$ is a set of all clusters. $r$ refers to the id of the cluster and $Clusters^{\left ( r \right )}$ refers to the $r$-th cluster.
The process of mapping nodes to clusters is shown in Figure \ref{fig4}.

\subsubsection{Bilevel Aggregation Module (BiAM)}\label{subsubsec3}

On the basis of the similarity clusters, we construct the cluster neighborhood $ S_{s}\left (o\right ) $, which is later used to aggregate and update the features of the target node. The cluster neighborhood $ S_{s}\left (o\right ) $ is specifically defined as follows:

\begin{equation}
S_{s}\left (o\right )= \left \{u\lvert u\in {V}\cap u \in Clusters^{\left ( r \right )} \right \}
\end{equation}

To ensure the permutation invariance of graph-structure data, we apply the bilevel aggregation scheme for the cluster neighborhood $ S_{s}\left (o\right ) $ to renew the characteristics of nodes. At the first level, the nodes in the same cluster are aggregated into a virtual node by means of an aggregation function. At the second level, we aggregate and update the virtual nodes aggregated in the first level together with the target node into the final node feature representation. The cluster is obtained by performing similarity mapping between the nodes in the graph and the target node. The similarity between nodes does not change with the order of the nodes in the graph, so the order of the nodes in the graph does not affect the clustering result. When the number of nodes in the cluster is constant, the mean aggregator satisfies permutation invariance. In addition, the result of first-level aggregation is the input of the second-level aggregation process and remains unchanged, thus making the entire bilevel aggregator satisfy permutation invariance. We utilize the mean aggregation function in the first-level aggregation step, and $\boldsymbol{e}_{(r)}^o$ is the final feature representation obtained after the first-level aggregation process, which is defined as follows:

\begin{equation}
\boldsymbol{e}_{(r)}^o =\frac{1}{\left \lvert {Clusters}^{(r)} \right \rvert } \sum_{u \in S_{s}(o)} \delta\left(\tau\left( u,o\right), {r}\right)\cdot \sigma^{(r)}(\boldsymbol{g}_u)
\end{equation}

\noindent
where $\lvert {Clusters}^{(r)} \rvert$ denotes the number of nodes that belong to the $r$-th cluster. $u$ is a node in the cluster neighborhood, and $\boldsymbol{g}_u$ is the value of node $u$. 

We define the linear transformation function $\sigma^{(r)}(\boldsymbol{x})$ as follows:

\begin{equation}
\sigma^{(r)}(\boldsymbol{x})= \left ( \boldsymbol{W}^{\left (  r\right )}  \boldsymbol{x}+\boldsymbol{b}^{\left ( r \right )}  \right )
\end{equation}

\noindent
where $\boldsymbol{W}^{\left (  r\right )}$ is the weight matrix and $\boldsymbol{x}$ is the feature representation of a node in the cluster neighborhood. $\boldsymbol{b}^{(r)}$ represents the bias vector.

\begin{figure}[t]
    \centering
    \includegraphics[width=9cm]{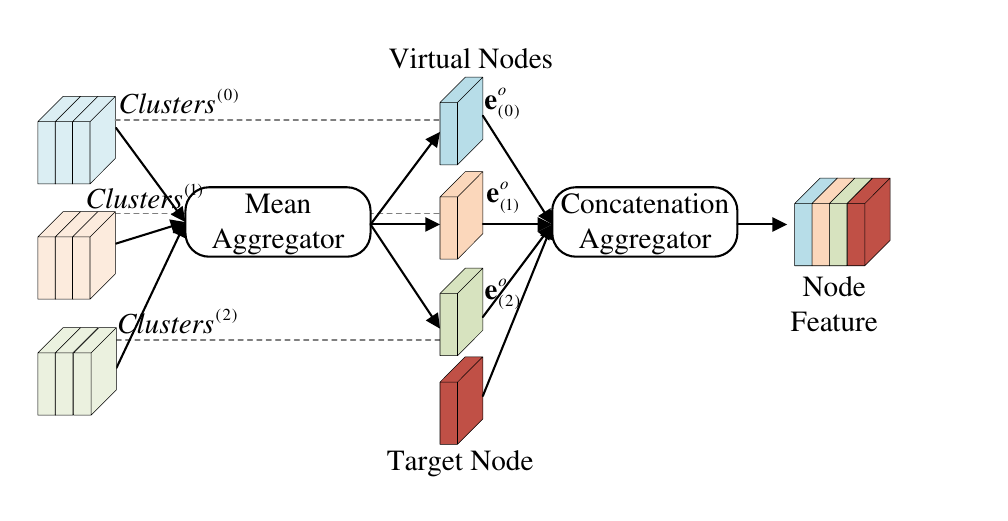}
    \\ \hspace*{\fill} \\
    \caption{\centering The detailed process of the bilevel aggregation polymerization.}
    \label{fig5}
\end{figure}

We specifically implement the $\delta\left(\tau\left(o,u\right), {r}\right)$ function as follows:

\begin{equation}
\delta\left(\tau\left( u,o\right), r\right)= \begin{cases}1, & \text { if } \tau( u,o)=  {r} \\  0, & \text { if }  \tau( u,o)\neq {r} \end{cases}
\end{equation}

\noindent
where $\delta\left ( \cdot, \cdot  \right )$ is the Kronecker delta function, which takes only nodes in the same cluster into account. This function is employed to separate different clusters for aggregation operations. The detailed process of the first-level aggregation is shown in Figure \ref{fig5}.

We perform further aggregation operations based on all virtual nodes $ \boldsymbol{e}_{(r)}^{o} $ and the target node. $\boldsymbol{h}_{i}$ is the final feature representation obtained after the second-level aggregator updates the target node. We define it as follows:

\begin{equation}
\boldsymbol{h}_{i}=\sigma\left({\boldsymbol{W}} \cdot (\boldsymbol{e}_{(r)}^o  \mathop{\lvert \rvert}\boldsymbol{g}_{i})\right) 
\end{equation}

\noindent
Here, we implement $\sigma\left ( \cdot  \right ) $ as a ReLU function. $i$ represents the $i$-th utterance in the conversation. $\boldsymbol{g}_{i}$ is the original feature representation of the target node updated only by Bi-LSTM, and $\lvert \rvert$ is the feature concatenation operation. $\boldsymbol{W} $ is the weight matrix of the node feature transformation. The detailed process of graph aggregation via bilevel aggregation according to the cluster neighborhood is shown in Figure \ref{fig5}.

\subsection{Emotion Classifier}\label{subsec3}

We take the updated feature representation of each node through the bilevel aggregated as the input for the final predicted label. Then, the following methods are used to predict the emotion labels  of the nodes:

\begin{equation}
\boldsymbol{l}_{i}=\sigma\left(\boldsymbol{W}_{l} \boldsymbol{h}_i+\boldsymbol{b}_{l}\right)
\end{equation}

\begin{equation}
\boldsymbol{p}_{i}=\operatorname{Softmax}\left(\boldsymbol{W}_{\text {smax }} \boldsymbol{l}_{i}+\boldsymbol{b}_\text {smax }\right)
\end{equation}

\noindent
where $\boldsymbol{h}_i$ represents the final feature of the target node, which contains multimodal information. $\boldsymbol{p}_{i}$ represents the probability vector of the emotion class for utterance $i$. $\boldsymbol{W}_{l}$, $\boldsymbol{b}_{l}$, $\boldsymbol{W}_{\text {smax} }$ and $\boldsymbol{b}_{\text {smax} } $ are all trainable parameters.

The category with the highest calculated probability is used as the prediction label, and the emotion label computation is defined as follows:

\begin{equation}
\hat {y}_{i} = \arg \max \limits (\boldsymbol{p}_{i})
\end{equation}

\subsection{Model Training}\label{subsec3}

We choose the categorical cross-entropy loss function during training. The calculation process is as follows:

\begin{equation}
\mathcal{L}=-\frac{1}{\sum_{i=1}^{K} N_{i}} \sum_{i=1}^{K} \sum_{j=1}^{N_{i}} \sum_{m=1}^{C} {y}_{i, j}^{(m)} \log \left({p}_{i, j}^{(m)}\right)
\end{equation}

\noindent
where $K$ is the number of conversations, and $m$ indicates the category of the emotion label. $y_{i, j}^{(m)}$ is the golden label for utterance $i$, ${p}_{i, j}^{(m)}$ is the predicted output for utterance $i$, and $N_i$ is the number of utterances in the conversation.

\section{Experimental Databases And Setup}

In this section, first, we introduce the datasets used for the experiments. Second, we describe details of the implementation of our method. Finally, we present the methodology for model evaluation and some state-of-the-art baselines.

\subsection{Datasets}\label{subsec4}

Both benchmark IEMOCAP \cite{7} and MELD \cite{8} datasets are used to measure the performance of RBA-GCN, and both contain three modalities: acoustic, visual and textual. Table \ref{tab1} presents  the detailed information of the two datasets, including the detailed distribution of each emotion and the number of utterances used for training, validation and testing. 

\noindent 
\textbf{IEMOCAP:} 
The University of Southern California has produced the IEMOCAP \cite{7} dataset. It contains up to 12 hours of multimodal audiovisual data, and there are 5 sessions in total, each consisting of a conversation between a man and a woman. The conversation is divided into two parts, namely, the fixed script and the free form, in a given thematic scene. The dataset has 151 conversations with a total of 7433 utterances  and is labeled with 6 types of emotions: “neutral”, “happy”, “sad”, “angry”, “frustrated” and “excited”, with non-neutral emotions accounting for 77$\%$.

\noindent
\textbf{MELD:} 
The MELD \cite{8} dataset is an extension of the EmotionLines Friends section of the plain text modality, and it is presented as a multiperson conversation, unlike the binary conversations in IEMOCAP. It contains 1433 conversations with 13708 utterances and is labeled with seven types of emotions, i.e., “anger”, “disgust”, “fear”, “joy”, “neutral”, “sadness”, and “surprise”, which are categorized into three categories, i.e., positive, negative and neutral, with nonneutral emotions accounting for 53\%.

\begin{table}[t]
\centering
\caption{\centering Statistics of the IEMOCAP dataset and the MELD dataset.}\label{tab1}

\begin{tabular}{|c|c|c|c|c|c|c|} 
\hline
\multirow{2}{*}{emotion} & \multicolumn{3}{c|}{IEMOCAP} & \multicolumn{3}{c|}{MELD}    \\ \cline{2-7} 

                  & train+val & test & sum       & train+val & test & sum                                           \\ 
\hline
Anger             & 869       & 234  & 1103      & 1262      & 345  & 1607    
   \\
Happiness/Joy     & 460       & 135  & 595       & 1906      & 402  & 2308                                         \\
Sadness           & 877       & 207  & 1084      & 794       & 208  & 1002                                         \\
Neutral           & 1387      & 321  & 1708      & 5180      & 1256 & 6436                                         \\
Excitement        & 828       & 213  & 1041      & --         & --    & --                                            \\
Frustration       & 1478      & 371  & 1849      & --         & --    & --                                            \\
Disgust           & --         & --    & --         & 293       & 68   & 361                                          \\
Surprise          & --         & --    & --         & 1355      & 281  & 1636                                         \\
Fear              & --       & --    & --         & 308       & 50   & 358                                          \\
\hline
\end{tabular}
\end{table}

\subsection{Data Preprocessing}\label{4.2}
During data preprocessing, TextCNN \cite{46} is utilized to extract raw textual features, the openSMILE toolkit with IS10 \cite{27} configuration is utilized to extract raw acoustic features, and DenseNet \cite{28} is utilized to extract raw visual facial expression features.

\subsection{Implementation Details}\label{4.3}

 In this subsection, we focus on the specific details of the RBA-GCN. We utilize the Adam optimizer to train the RBA-GCN. 
 The model is configured with a dropout rate of 0.5, a $\rho$ parameter of 0.3, a learning rate of 0.0009, and a $\gamma$ parameter of 8. The model is trained for up to 1500 epochs.

\subsection{Evaluation Metrics}\label{4.4}
As shown in Table \ref{tab1}, there are inherent data imbalances in the IEMOCAP and MELD datasets. Considering that the weighted average F1 score has good ability to handle unbalanced classes, in the following experiments, we employ it as our metric for the evaluation of our proposed RBA-GCN. The weighted average F1 is shown below:

\begin{equation}
W A F 1=\frac{\sum_{j=1}^{M} N_{j} \cdot F 1_{j}}{\sum_{j=1}^{M} N_{j}}
\end{equation}

\noindent
where the number of emotion categories in the dataset is denoted by $M$ and the sample size of a category is denoted by $N_j$. $F1_j$ is the $f1$ score of samples in a category.

\subsection{State-of-the-art Baselines}\label{4.5}

In this subsection, we present several of the most advanced baseline methods. To highlight the superiority of RBA-GCN, we compare our proposed method with these baselines.

\noindent
\textbf{BC-LSTM} \cite{49}: A context-aware utterance representation for emotion classification is utilized, and the model aims to capture contextual information through a Bi-LSTM layer.

\noindent
\textbf{CMN} \cite{1}: The context of a particular speaker is modeled by the different memories of each speaker, and the historical utterance of each speaker is modeled separately by GRU as a memory unit.

\noindent
\textbf{ICON} \cite{2}: The attention mechanism is used to obtain the result fusion of memory units with the current utterance representation for utterance emotion classification.

\noindent
\textbf{GAT} \cite{54}: The model applies an attention mechanism to characterize the importance of neighboring nodes to nodes and updates node features for emotion recognition using different edge weights.

\noindent
\textbf{DialogueRNN} \cite{3}: To capture speaker information, the context of previous utterances and affective information, three types of states, namely, speaker state, global state, and emotional state, are employed.

\noindent
\textbf{DialogueGCN} \cite{4}: This is the first time that GCNs are applied to an emotion recognition scenario in a conversation. It can effectively model the contextual information and speaker information in a conversation.

\noindent
\textbf{MTAG} \cite{55}: This method converts unaligned multimodal sequence data into a graph with heterogeneous nodes and edges to capture the rich interactions across modalities and through time.

\noindent
\textbf{ConGCN} \cite{48}: This method constructs the entire dataset as a graph and uses subgraphs in the larger graph to represent each conversation. Speaker nodes are also connected to corresponding utterance nodes, which are used to model speaker-sensitive dependencies.

\noindent
\textbf{MMGCN} \cite{5}: This approach utilizes multimodal information based on DialogueGCN. The model uses spectral domain GCNs to encode the multimodal graph, which makes it possible for multilayer GCNs to capture more distant contextual information. However, it does not consider the relationships between nodes in the graph.

\section{Results and Discussions}
In this section, first, we compare our proposed method with all the baseline methods mentioned in Subsection \ref{4.5} to verify the superiority of our approach. Second, we perform a case study to further validate our approach.
Then, we evaluate the effectiveness of the three modules in RBA-GCN.
Finally, we explore the importance of effectively capturing the interactions between different models.

\begin{table*}[]
\centering
\caption{\centering Comparison with baseline methods on the IEMOCAP dataset.}\label{tab3}
\begin{tabular}{|c|cccccc|c|}
\hline
\multirow{2}{*}{Models} & \multicolumn{7}{c|}{IEMOCAP}                                                                                                                                                           \\  \cline{2-8}  
                  & \multicolumn{1}{c|}{happy} & \multicolumn{1}{l|}{sad}   & \multicolumn{1}{c|}{neutral} & \multicolumn{1}{c|}{angry} & \multicolumn{1}{c|}{excited} & \multicolumn{1}{c|}{frustrated} & WAF1 \\ \hline
BC-LSTM \cite{49}               & \multicolumn{1}{c|}{0.3443} & \multicolumn{1}{c|}{0.6087} & \multicolumn{1}{c|}{0.5181}   & \multicolumn{1}{c|}{0.5673} & \multicolumn{1}{c|}{0.5795}   & \multicolumn{1}{c|}{0.5892}      & 0.5495      \\ \hline
CMN \cite{1}               & \multicolumn{1}{c|}{0.3038} & \multicolumn{1}{c|}{0.6241} & \multicolumn{1}{c|}{0.5239}   & \multicolumn{1}{c|}{0.5983} & \multicolumn{1}{c|}{0.6025}   & \multicolumn{1}{c|}{0.6069}      & 0.5613      \\ \hline
ICON \cite{2}             & \multicolumn{1}{c|}{0.2991} & \multicolumn{1}{c|}{0.6457} & \multicolumn{1}{c|}{0.5738}   & \multicolumn{1}{c|}{0.6304} & \multicolumn{1}{c|}{0.6342}   & \multicolumn{1}{c|}{0.6081}      & 0.5854      \\ \hline
DialogueRNN \cite{3}      & \multicolumn{1}{c|}{0.3318} & \multicolumn{1}{c|}{0.7880}  & \multicolumn{1}{c|}{0.5921}   & \multicolumn{1}{c|}{0.6528} & \multicolumn{1}{c|}{0.7186}   & \multicolumn{1}{c|}{0.5891}      & 0.6275      \\ \hline
DialogueGCN \cite{4}      & \multicolumn{1}{c|}{0.4275}  & \multicolumn{1}{c|}{0.8088} & \multicolumn{1}{c|}{0.5871}   & \multicolumn{1}{c|}{0.6608} & \multicolumn{1}{c|}{0.6997}   & \multicolumn{1}{c|}{0.6121}      & 0.6418      \\ \hline
GAT \cite{54}      & \multicolumn{1}{c|}{0.4761}  & \multicolumn{1}{c|}{0.6962} & \multicolumn{1}{c|}{0.5869}   & \multicolumn{1}{c|}{0.6428} & \multicolumn{1}{c|}{0.6750}   & \multicolumn{1}{c|}{0.5857}      & 0.6367      \\ \hline
GAT-fully \cite{54}      & \multicolumn{1}{c|}{0.4720}  & \multicolumn{1}{c|}{0.7343} & \multicolumn{1}{c|}{0.6052}   & \multicolumn{1}{c|}{0.6523} & \multicolumn{1}{c|}{0.6638}   & \multicolumn{1}{c|}{0.5603}      & 0.6516      \\ \hline
MTAG \cite{55}      & \multicolumn{1}{c|}{0.3603}  & \multicolumn{1}{c|}{0.7136} & \multicolumn{1}{c|}{0.5051}   & \multicolumn{1}{c|}{0.4836} & \multicolumn{1}{c|}{0.6030}   & \multicolumn{1}{c|}{0.5579}      & 0.5533      \\ \hline
MMGCN \cite{5}            & \multicolumn{1}{c|}{0.4234} & \multicolumn{1}{c|}{0.7867} & \multicolumn{1}{c|}{0.6173}   & \multicolumn{1}{c|}{\textbf{0.6900}}    & \multicolumn{1}{c|}{\textbf{0.7433}}   & \multicolumn{1}{c|}{0.6232}      & 0.6622      \\ \hline
Ours              & \multicolumn{1}{c|} {\textbf{0.7166}} & \multicolumn{1}{c|}{\textbf{0.8695}} & \multicolumn{1}{c|}{\textbf{0.6768}}   & \multicolumn{1}{c|}{0.6666} & \multicolumn{1}{c|}{0.6800}   & \multicolumn{1}{c|}{\textbf{0.6950}}      & \textbf{0.7143}      \\ \hline
\end{tabular}
\end{table*}

\begin{table*}[]
\centering
\caption{\centering Comparison with baseline methods on the MELD dataset.}\label{tab4}
\begin{tabular}{|c|ccccccc|c|}
\hline
\multirow{2}{*}{Models} & \multicolumn{7}{c|}{MELD}                                                                                                                                                                   &            \\ \cline{2-9}
                  & \multicolumn{1}{c|}{anger} & \multicolumn{1}{c|}{disgust} & \multicolumn{1}{c|}{fear} & \multicolumn{1}{c|}{joy}   & \multicolumn{1}{c|}{neutral} & \multicolumn{1}{c|}{sadness} & \multicolumn{1}{c|}{surprise} & WAF1 \\ \hline
BC-LSTM \cite{49}             & \multicolumn{1}{c|}{0.445}  & \multicolumn{1}{c|}{0}       & \multicolumn{1}{c|}{0}    & \multicolumn{1}{c|}{0.497}  & \multicolumn{1}{c|}{0.764}    & \multicolumn{1}{c|}{0.156}    & 0.484     & 0.568       \\ \hline
CMN \cite{1}              & \multicolumn{1}{c|}{0.447}  & \multicolumn{1}{c|}{0}       & \multicolumn{1}{c|}{0}    & \multicolumn{1}{c|}{0.477}  & \multicolumn{1}{c|}{0.743}    & \multicolumn{1}{c|}{0.234}    & 0.472     & 0.559       \\ \hline
ICON \cite{2}             & \multicolumn{1}{c|}{0.448}  & \multicolumn{1}{c|}{0}       & \multicolumn{1}{c|}{0}    & \multicolumn{1}{c|}{0.502}  & \multicolumn{1}{c|}{0.736}    & \multicolumn{1}{c|}{0.232}    & 0.500       & 0.563       \\ \hline
GAT \cite{54}      & \multicolumn{1}{c|}{0.4262}  & \multicolumn{1}{c|}{0}     & \multicolumn{1}{c|}{0}  & \multicolumn{1}{c|}{0.5128}  & \multicolumn{1}{c|}{0.6154}    & \multicolumn{1}{c|}{0.2307}    & 0.4182     & 0.5045     \\ \hline
GAT-fully \cite{54}      & \multicolumn{1}{c|}{0.4323}  & \multicolumn{1}{c|}{0}     & \multicolumn{1}{c|}{0}  & \multicolumn{1}{c|}{0.5186}  & \multicolumn{1}{c|}{0.6363}    & \multicolumn{1}{c|}{0.2197}    & 0.4256     & 0.5166     \\ \hline
MTAG \cite{55}      & \multicolumn{1}{c|}{0.4742}  & \multicolumn{1}{c|}{0}     & \multicolumn{1}{c|}{0}  & \multicolumn{1}{c|}{0.5361}  & \multicolumn{1}{c|}{0.7002}    & \multicolumn{1}{c|}{0.2464}    & 0.4793     & 0.5824      \\ \hline
DialogueRNN \cite{3}      & \multicolumn{1}{c|}{0.415}  & \multicolumn{1}{c|}{0.017}     & \multicolumn{1}{c|}{0.012}  & \multicolumn{1}{c|}{0.507}  & \multicolumn{1}{c|}{0.735}    & \multicolumn{1}{c|}{0.238}    & 0.494     & 0.5711      \\ \hline
ConGCN \cite{48}       & \multicolumn{1}{c|}{0.468}  & \multicolumn{1}{c|}{0.106}    & \multicolumn{1}{c|}{\textbf{0.087}}  & \multicolumn{1}{c|}{0.531}  & \multicolumn{1}{c|}{\textbf{0.767}}    & \multicolumn{1}{c|}{0.285}    & 0.503     & 0.5823      \\ \hline
Ours              & \multicolumn{1}{c|}{\textbf{0.5000}}    & \multicolumn{1}{c|}{\textbf{0.1132}}   & \multicolumn{1}{c|}{0.0752} & \multicolumn{1}{c|}{\textbf{0.5714}} & \multicolumn{1}{c|}{0.7143}   & \multicolumn{1}{c|}{\textbf{0.3333}}   & \textbf{0.5556}     & \textbf{0.6267}      \\ \hline
\end{tabular}
\end{table*}

\subsection{Comparison with State-of-the-art Baselines}\label{subsec5}
We compare the RBA-GCN with the baseline methods on IEMOCAP and MELD in Subsection \ref{4.5}. Table \ref{tab3} and Table \ref{tab4} show the comparison results.
The experimental results indicate that our method significantly outperforms all the baseline methods. On the IEMOCAP dataset, the RBA-GCN achieves a WAF1 score of 71.43\%, which is 5 points higher than that of the most advanced existing method. In addition, it achieves a WAF1 score of 62.67\% on the MELD dataset, which is 4 points higher than that of the best baseline method. Furthermore, we compare the proposed approach with the GAT and MTAG  models. Unlike the graph attention mechanism, which automatically removes edges with low weights or directly assigns low weights during aggregation, RBA-GCN employs similarity measures to filter out redundant information and map nodes to different clusters. Finally, intracluster aggregation and intercluster aggregation are performed. We conduct additional experiments to compare the performance of different graph generation methods on the GAT. “GAT” involves using our graph generation method, and “GAT-fully” represents the fully graph connected method. The experimental results are shown in Tables \ref{tab3} and \ref{tab4}. The overall performance of GAT-fully is better than that of the GAT. This is because GAT-fully is better than GAT at capturing contextual information. The comparison results show that our proposed RBA-GCN performs better than both the GAT and GAT-fully, which indicates the superiority of the RBA-GCN.

The experimental results for the IEMOCAP dataset are shown in Table \ref{tab3}. These results demonstrate that our method obtains the best scores on almost all the labels. Undoubtedly, our method achieves the most advanced weighted average F1 score. Because the IEMOCAP dataset has more than 70 conversations and the average conversation length exceeds 50 utterances, DialogueGCN and MMGCN use sliding windows in the composition to reduce the complexity of the graph. Although this approach reduces the complexity of the model, it loses the context dependency of the target node over long ranges.  
Remarkably, our RBA-GCN method achieves the best prediction result, with a WAF1 score of 71.66\% for the “happy” label, which is almost 30\% higher than that of the best performing DialogueGCN model on this label. Data with a “happy” label in the IEMOCAP dataset account for only 7\% of the whole dataset. This means that the probability of “happy” appearing in a conversation is minimal and a correct prediction for such an utterance is difficult. As a result, the prediction accuracy of such labels is very low. The DialogueGCN model uses GCNs to aggregate node information, which improves the prediction accuracy of such nodes. Due to the complexity of graph generation, the prediction of MMGCN for such node labels is not satisfactory. In our method, for such nodes, we first filter the noisy information using clusters to reduce the redundancy of the target node information. Then, we enhance the favorable features in each cluster to improve the classification effect. Finally, bilevel aggregation effectively captures the long-range contextual information and makes excellent use of the interaction between multiple modalities, thereby improving the prediction accuracy of such nodes more effectively.

The experimental results of the MELD dataset are displayed in Table \ref{tab4}. These results indicate that our method achieves the optimal scores on almost every label. The MELD dataset consists of multiperson conversations, which are briefer and have few specific emotional expressions compared to the IEMOCAP dataset. In addition, the average conversation length of the MELD dataset is more than 10 utterances. 
Since there are more than 4 speakers in many conversations, only a few utterances are available for most speakers in a conversation. 
These factors make it more difficult to improve the classification accuracy. However, the prediction results of our model on the MELD dataset are also improved by at least 4 points compared to that of ConGCN. The substantial improvement is due to our cluster and bilevel aggregation approach.

\begin{figure*}[]%
\centering
\includegraphics[width=0.8\textwidth]{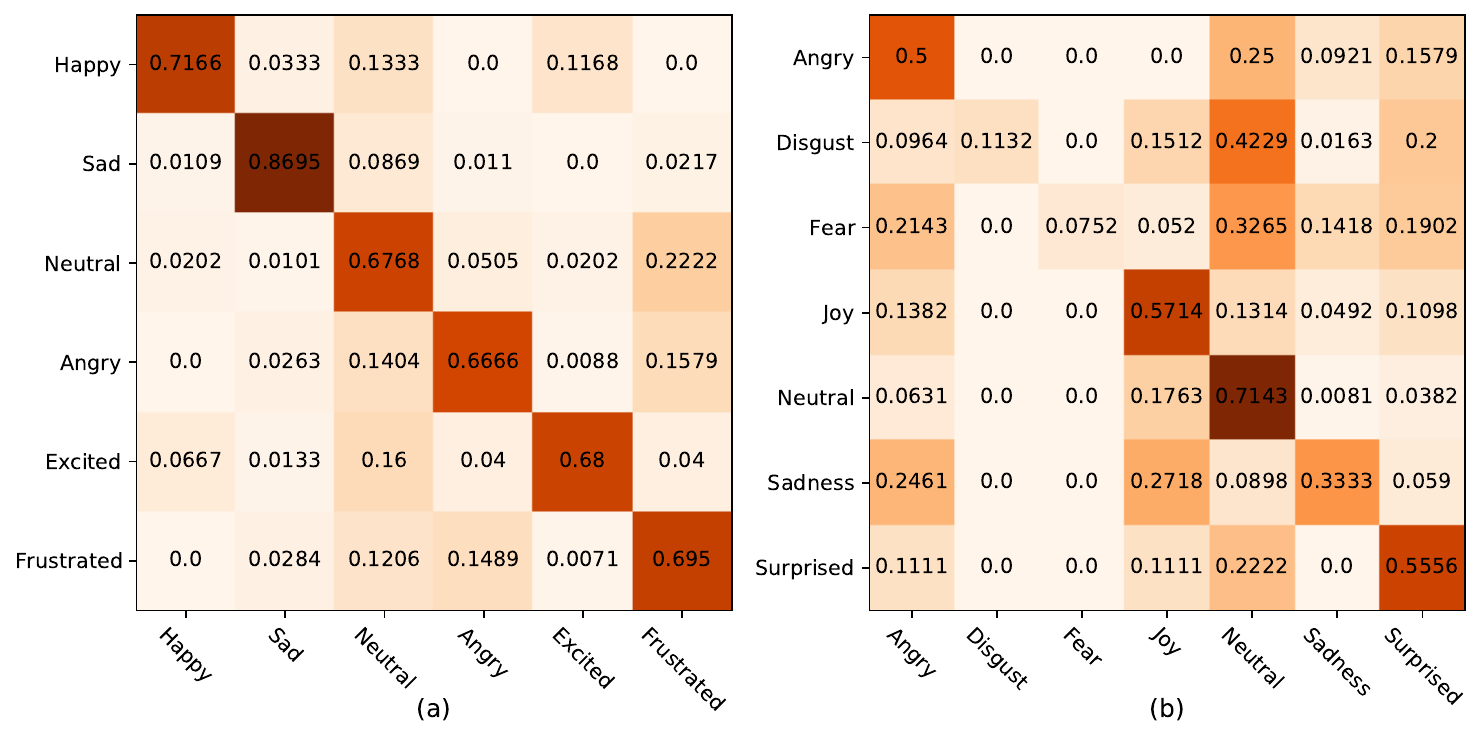}
\caption{\centering Confusion matrix of proposed RBA-GCN on: (a) IEMOCAP dataset, and (b) MELD dataset. Note: x-axis is the correct label, y-axis is the predicted label.  }\label{fig6}
\end{figure*}

The confusion matrix of our RBA-GCN method with respect to the IEMOCAP and MELD datasets is shown in Figure \ref{fig6}, which illustrates the effectiveness of our method more distinctly. For the IEMOCAP dataset,  the weighted average F1 scores of all classes are relatively balanced, with the “sad” category having the highest weighted average F1 score of 86.95\%. For the MELD dataset, we find from Table \ref{tab1} that the training and test sets for the three categories of “disgust”, “fear” and “sadness” are relatively small compared to those of other categories. Although the MELD dataset has obvious class imbalance, leading to more difficulty in model training, RBA-GCN is significantly improved. Thus, RBA-GCN can filter out noisy information, reduce the redundancy of target node information, and better retain node discriminant information. 
Additionally, our model can effectively capture long-range contextual information and interactions between modalities.

\subsection{RBA-GCN under Various Modality Settings}

We experimentally compare the performance between single-modality and multimodality settings to verify the effectiveness of RBA-GCN for multimodal interactions. The performance of our proposed method in various modality settings is shown in Table \ref{tab6}.
 
According to the results in Table \ref{tab6}, there are some differences in the performance of each modality under the single-modality setting, with the textual modality performing best. We argue that textual features can express emotions more intuitively than acoustic and visual features in a conversational emotion recognition task. With few exceptions, the words for emotional expression are in the utterance.

In a multimodal setting, the performance of multiple-modality fusion is better than that of individual modalities, but the best performance is obtained with the fusion of three modalities. We believe that multiple modality features can complement each other compared to a single modality. Similar to communication with people in reality, we can combine facial expressions, voice and conversation content to determine mood fluctuations of the speaker.
The experimental results indicate that the RBA-GCN achieves a significant improvement on most modal combinations compared to the multimodal fusion method from MMGCN. This indicates that our multimodal fusion method can fuse sufficient information effectively. Meanwhile, the node discriminant information can be retained after multimodal fusion, which makes the emotion recognition more accurate.

\begin{figure*}[t]%
\centering
\includegraphics[width=0.8\textwidth]{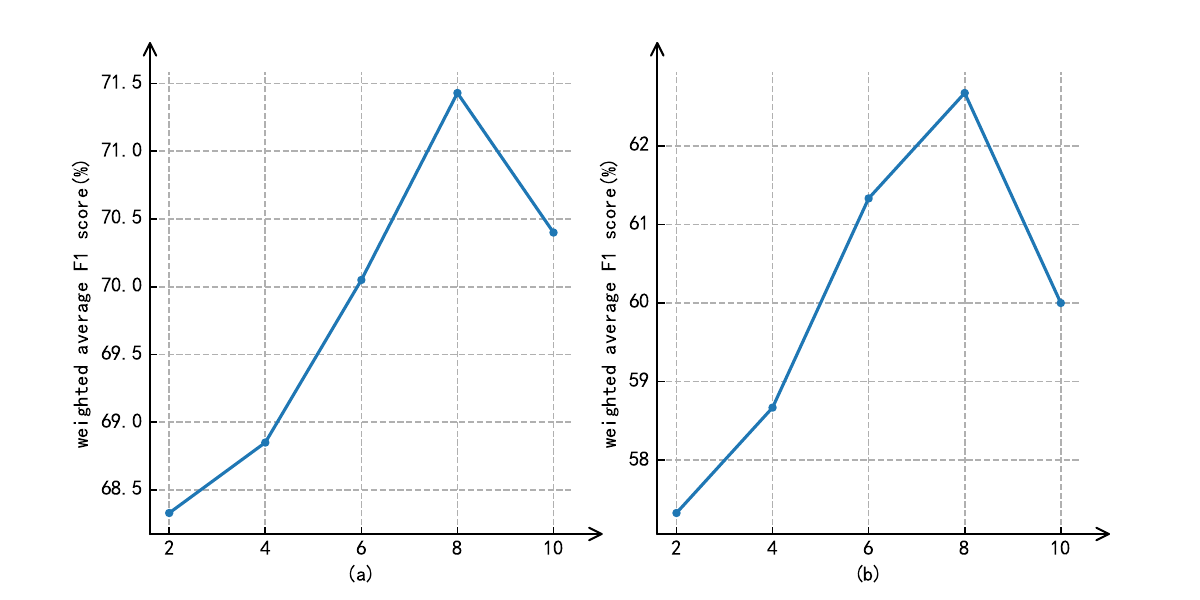}
\caption{\centering (a) Representation of different quantitative clusters on the IEMOCAP dataset; (b) Representation of different quantitative clusters on the MELD dataset. Note: the horizontal axis represents the value of $\gamma$.}\label{fig7}
\end{figure*}

\begin{table*}[h]
\centering
\caption{\centering Performance comparison with different modalities on the IEMOCAP dataset. Note: T=Text, A=Audio, V=Video.}\label{tab6}
\begin{tabular}{|c|ccccccc|}
\hline
\multirow{2}{*}{Modality} & \multicolumn{7}{c|}{IEMOCAP}                                                                                                                                                                      \\  \cline{2-8}
                  & \multicolumn{1}{c|}{happy} & \multicolumn{1}{l|}{sad}   & \multicolumn{1}{c|}{neutral} & \multicolumn{1}{c|}{angry} & \multicolumn{1}{c|}{excited} & \multicolumn{1}{c|}{frustrated} & WAF1 \\ \hline
T      & \multicolumn{1}{c|}{0.7090}  & \multicolumn{1}{c|}{0.8366} & \multicolumn{1}{c|}{0.6363}   & \multicolumn{1}{c|}{0.5522} & \multicolumn{1}{c|}{0.6455}   & \multicolumn{1}{c|}{0.6461}      & 0.6609     \\ \hline
V       & \multicolumn{1}{c|}{0.6326} & \multicolumn{1}{c|}{0.5572} & \multicolumn{1}{c|}{0.4473}   & \multicolumn{1}{c|}{0.4385} & \multicolumn{1}{c|}{0.5769}   & \multicolumn{1}{c|}{0.5052}      & 0.5129      \\ \hline
A       & \multicolumn{1}{c|}{0.6034} & \multicolumn{1}{c|}{0.7352} & \multicolumn{1}{c|}{0.4536}   & \multicolumn{1}{c|}{0.5200}    & \multicolumn{1}{c|}{0.5875}   & \multicolumn{1}{c|}{0.5882}      & 0.5783      \\ \hline
T+V     & \multicolumn{1}{c|}{0.7049} & \multicolumn{1}{c|}{0.8421} & \multicolumn{1}{c|}{\textbf{0.7032}}   & \multicolumn{1}{c|}{0.6239} & \multicolumn{1}{c|}{0.6410}    & \multicolumn{1}{c|}{0.6906}      & 0.6988      \\ \hline
T+A     & \multicolumn{1}{c|}{\textbf{0.7288}} & \multicolumn{1}{c|}{0.8404} & \multicolumn{1}{c|}{0.6200}      & \multicolumn{1}{c|}{0.5932} & \multicolumn{1}{c|}{\textbf{0.6973}}   & \multicolumn{1}{c|}{0.6764}      & 0.6850       \\ \hline
V+A     & \multicolumn{1}{c|}{0.6724} & \multicolumn{1}{c|}{0.7428} & \multicolumn{1}{c|}{0.4695}   & \multicolumn{1}{c|}{0.5833} & \multicolumn{1}{c|}{0.6714}   & \multicolumn{1}{c|}{0.6194}      & 0.6162      \\ \hline
T+V+A   & \multicolumn{1}{c|}{0.7166} & \multicolumn{1}{c|}{\textbf{0.8695}} & \multicolumn{1}{c|}{0.6768}   & \multicolumn{1}{c|}{\textbf{0.6666}} & \multicolumn{1}{c|}{0.6800}      & \multicolumn{1}{c|}{\textbf{0.6950}}       & \textbf{0.7143}      \\ \hline

\end{tabular}
\end{table*}

\subsection{Comparison with Other Fusion Methods}

A performance comparison of RBA-GCN and the most advanced baseline methods is shown in Table \ref{tab8}. We compare our proposed method with other multimodal fusion methods, including other representative fusion methods such as MFN, MMGCN and CTNet, to illustrate the superiority of the RBA-GCN.

\begin{table}[h]
\centering
\caption{\centering Performance comparison  with advanced multimodal fusion methods on the IEMOCAP dataset and MELD dataset.}\label{tab8}
\begin{tabular}{c|c|c}
\hline
Multimodal   fusion method & IEMOCAP & MELD \\ \hline
MFN \cite{50}                       & 0.6277 & 0.5470   \\
MulT  \cite{51}                     & 0.6237 & 0.5649   \\
MMGCN  \cite{5}                    & 0.6622  & 0.5865  \\
CTNet   \cite{47}                   & 0.6750  & 0.6050  \\
Ours                        & \textbf{0.7143} & \textbf{0.6267}  \\ \hline
\end{tabular}
\end{table}

Our method outperforms other multimodal fusion methods on both datasets, as shown in Table \ref{tab8}. On the IEMOCAP dataset, it outperforms the most advanced graph convolution fusion method (MMGCN) by more than 5\% and is nearly 4\% higher than the current most advanced fusion method (CTNet). On the MELD dataset, our method outperforms the most advanced graph convolution fusion method (MMGCN) by more than 4\% and is nearly 2\% higher than the most advanced fusion method (CTNet). This reflects the superiority of our proposed multimodal fusion method anchored on relational bilevel aggregation, which can effectively capture the interactions between modalities.

\subsection{Ablation Study}\label{subsec5}

Ablation studies are conducted to demonstrate the effectiveness of the various components of our proposed method (RBA-GCN).

\subsubsection{The effectiveness of the graph generation module (GGM)}\label{subsec5}

To verify the effectiveness of our GGM, we compare the experimental results of our method with those of previous methods for graph generation. We compare the graph generation of the fully connected graph with our graph generation method while ensuring that other conditions of the network structure remain unchanged. The experimental results are shown in Table \ref{tab9}.

\begin{table}[h]
\centering
\caption{\centering Performance comparison with other graph generation methods on the IEMOCAP and MELD datasets.}\label{tab9}
\begin{tabular}{c|c|c}
\hline
Graph generation method & IEMOCAP & MELD      \\ \hline
Fully connected graph     & 0.7057   &   0.5733         \\
Our graph generation      & \textbf{0.7143} & \textbf{0.6267}\\ \hline
\end{tabular}
\end{table}

Our graph generation method performs better on the IEMOCAP dataset than other methods with fully connected graphs by nearly 1\%. On the MELD dataset, our graph generation method outperforms the fully connected graph generation method employed by other models by 5\%. To explain the superior performance of our graph generation method on the MELD dataset, we argue that only a small number of utterances per conversation by most participants in this dataset lead to increased information redundancy. However, our graph generation method can effectively reduce the information redundancy of the target node and retain the discriminant information of the node. This leads to a significant improvement in the experimental results.

\subsubsection{Effectiveness of the similarity-based cluster building module (SCBM) }\label{subsec5}

To verify the effectiveness of our clusters and demonstrate that the clusters can effectively filter information irrelevant to the target node, an ablation study is performed. According to the data in Table \ref{tab10}, the clusters significantly influence the final classification result of the model. Consequently, irrelevant information can be effectively filtered so that the discriminant information of the target node is better retained.

We further compare the performance with and without clusters under different combinations of modalities. As shown in Table \ref{tab10}, the performance with clusters is better than that without clusters for different combinations of modalities. This demonstrates the effectiveness of clusters for multimodal interactions.

\begin{table}[h]
\centering
\caption{\centering  The impact of clusters on ERC performance. Note: T=Text, A=Audio, V=Video.}\label{tab10}
\begin{tabular}{|c|c|c|c|}
\hline
\multicolumn{1}{|l|}{RBA-GCN}      & Modalities   & IEMOCAP         & MELD            \\ \hline
\multirow{4}{*}{w/o Clusters} & A+T   & 0.6265          & 0.5200            \\ \cline{2-4} 
                                   & A+V   & 0.6076          & 0.4933          \\ \cline{2-4} 
                                   & T+V   & 0.6368          & 0.5333          \\ \cline{2-4} 
                                   & A+T+V & 0.6489          & 0.5467          \\ \hline
\multirow{4}{*}{w Clusters}   & A+T   & 0.6850           & 0.5357          \\ \cline{2-4} 
                                   & A+V   & 0.6162          & 0.5067          \\ \cline{2-4} 
                                   & T+V   & 0.6988          & 0.5779          \\ \cline{2-4} 
                                   & A+T+V & \textbf{0.7143} & \textbf{0.6267} \\ \hline
\end{tabular}
\end{table}

The number of clusters $\gamma$ is a key hyperparameter for bilevel aggregation. Intuitively, the final classification performance of RBA-GCN is influenced by the value of $\gamma$. Our aggregation uses clusters to perform the first-level aggregation operation because the value of $\gamma$ affects the cluster number. Therefore, to study the effect of clusters on model performance, we choose $\gamma$ in $\left \{2, 4, 6, 8, 10 \right \} $.

Our experiments show that with increasing cluster number within a certain range, a consistent improvement is observed. As shown in Figure \ref{fig7}, RBA-GCN has the best classification performance when $\gamma=8$. The weighted average F1 scores are 71.43\% and 62.67\% on the IEMOCAP and MELD datasets, respectively. The increase in the cluster number allows for more detailed differentiation of other nodes so that similar nodes can be better aggregated. However, the model classification performance decreases when $\gamma > 8$. We 
believe 
that the number of clusters impacts the second-level aggregation. The greater the number of clusters is, the more virtual nodes are aggregated, which destroys the target node information.

To further investigate the effectiveness of the SCBM, a finer-grained ablation study is performed. The experimental results in Table \ref{tab11} show that our method achieves the best results. This proves the effectiveness of our application of connected neighborhood $C_{g}(o)$ to retain multimodal information and disconnected neighborhood $D_{g}(o)$ with filter $s(u,o)$ to retain long-range contextual information.

\begin{table}[h]
\centering
\caption{\centering  Ablation study on IEMOCAP dataset. }\label{tab11}
\begin{tabular}{|c|c|c|c|}
\hline
RBA-GCN                   & WAF1 \\ \hline
$C_{g}(o)$                     & 0.6024  \\ \hline
$D_{g}(o)$                    & 0.6093  \\ \hline
$C_{g}(o) \; with \; s(u,o)$             & 0.5921  \\ \hline
$D_{g}(o) \; with \; s(u,o)$             & 0.6231  \\ \hline
$C_{g}(o) + D_{g}(o)$              & 0.6813  \\ \hline
$C_{g}(o) \; with \; s(u,o)+D_{g}(o)$          & 0.6489  \\ \hline
$C_{g}(o) \; with \; s(u,o)+D_{g}(o) \; with \; s(u,o)$    & 0.6895 \\ \hline
$C_{g}(o)+D_{g}(o) \; with \; s(u,o)$ \textbf{(ours)}  & \textbf{0.7143}  \\ \hline
\end{tabular}
\end{table}

\subsubsection{Effect of GCN layers on RBA-GCN }\label{subsec5}

\begin{figure}[]
    \centering
    \includegraphics[width=8cm]{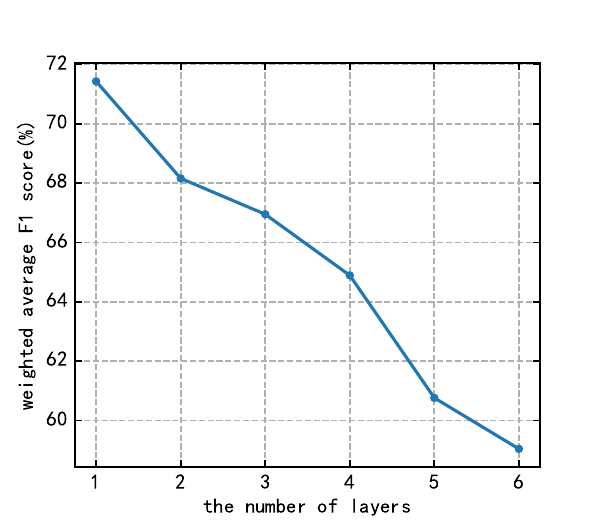}
    \\ \hspace*{\fill} \\
    \caption{\centering  Performance with the different number of  GCN layers on RBA-GCN on IEMOCAP. }
    \label{layer}
\end{figure}

We conduct a comparison study on the number of GCN layers on RBA-GCN. The experimental results are shown in Figure \ref{layer}, where the best results are achieved when we apply only bilevel aggregation. The model performance begins to degrade when we scale up the number of GCN layers. This further validates that after multilayer GCNs aggregation, the nodes in the graph become very similar and may lose the discriminant information of the node.

\subsection{Complexity Analysis}\label{subsec5}

\begin{figure}[t]
    \centering
    \includegraphics[width=8cm]{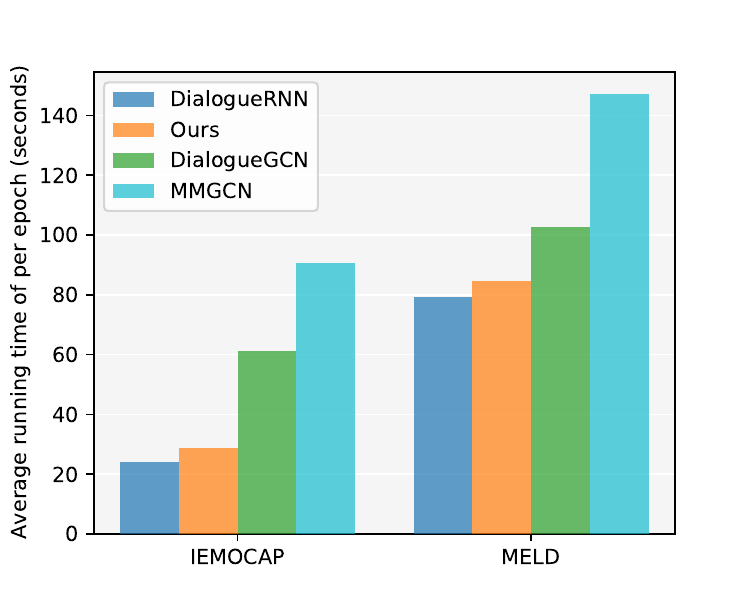}
    \\ \hspace*{\fill} \\
    \caption{\centering  Running time comparison of four models. }
    \label{fig8}
\end{figure}

The temporal complexity of GCNs is very important because certain conversations in the real world tend to be relatively long. Therefore, the graphs composed of these conversations are very large and have a very complex structure. In this subsection, we compare the temporal complexity of our method with that of the methods in Section \ref{4.5}. We compare the actual runtime (1500 epochs) of the DialogueGCN, MMGCN, DialogueRNN and RBA-GCN models on all datasets using the hyperparameters described in Section \ref{4.3}. According to the data in Figure \ref{fig8}, DialogueRNN takes the least time, while our method comes in second place.
We believe that our model is computationally complex and tedious compared to traditional neural network models, which is a significant reason why it consumes more time. Next is DialogueGCN, and MMGCN is the slowest. Although these methods employ some computational optimization techniques, such as sliding windows, they have not significantly reduced their computational cost. Due to the tremendous number of conversations in real life every day, the graph is large. Therefore, in future work, we will consider how to reduce the training time and enhance the robustness of the model.

\subsection{Case Study}

For a more intuitive comparative analysis of our method and more advanced methods, we perform a case study. Table \ref{tab5} shows the results of our analysis for one case on the IEMOCAP dataset, where the results in red indicate incorrect predictions and the results in green indicate correct predictions. According to the prediction results, our method clearly outperforms the other methods. We think that most of the utterances in this conversation are “neutral”, while some other emotion-labeled utterances are mixed into the conversation. Since traditional graph convolution methods aggregate the information of neighboring nodes, this leads to target node discriminant information loss and prediction errors. In this case, the prediction results of other methods for these utterances are wrong, while our model handles these cases well. In particular, the fifth utterance is predicted as “neutral” by other models, while our model produces the correct label “happy”. This is because our method can effectively capture long-range contextual information and interactions between modalities by considering the relevance between nodes and filtering out noisy information.

\begin{table*}[]
\centering
\caption{\centering A case study on the IEMOCAP dataset. RED lettering INDICATES WRONG PREDICTION, GREEN lettering INDICATES CORRECT PREDICTION. 
 }\label{tab5}
\resizebox{\textwidth}{!}{
\begin{tabular}{|c|l|c|c|c|c|c|}
\hline
Turn & \multicolumn{1}{c|}{Utterances}                                                                                                                             & Label   & DialogueRNN \cite{3}                    & DialogueGCN \cite{4}                    & MMGCN \cite{5}                         & Ours                            \\ \hline
1    & \begin{tabular}[c]{@{}l@{}}\textbf{A: }Did we bring something less? You forgot to bring \\ the baby's anvil?\end{tabular} & neutral & neutral                        & neutral                        & neutral                        & neutral                        \\ 
2    & \multicolumn{1}{r|}{\textbf{B:} Women like babies it's common knowledge, okay?}                                                            & neutral & neutral                        & neutral                        & neutral                        & neutral                        \\ 
3    & \multicolumn{1}{r|}{\textbf{B: }Women like men who like babies.}                                                                                & neutral & neutral                        & neutral                        & neutral                        & neutral                        \\ 
4    & \multicolumn{1}{r|}{\textbf{B: }Quick, point him towards that group of beautiful women.}                                                                & neutral & neutral                        & neutral                        & neutral                        & neutral                        \\ 
5    & \multicolumn{1}{r|}{\textbf{B: }No, no, wait, to get them, we got one, on the left.}                                                                & happy   & \textcolor{red}{neutral} & \textcolor{red}{neutral} & \textcolor{red}{neutral} & \textcolor[rgb]{0,0.502,0}{happy}   \\ 
6    & \multicolumn{1}{r|}{\textbf{B: }Well, give me the baby.}                                                                                       & neutral & neutral                        & neutral                        & neutral                        & neutral                        \\ 
7    & \textbf{A: }No, I got him.                                                                                                                       & neutral & neutral                        & neutral                        & neutral                        & neutral                        \\ 
8    & \textbf{A: }Oh, you really wanted him?                                                                                                         & excited &  \textcolor{red}{neutral} & \textcolor[rgb]{0,0.502,0} {excited}                 & \textcolor[rgb]{0,0.502,0} {excited}                        & \textcolor[rgb]{0,0.502,0} {excited} \\ 
9    & \multicolumn{1}{r|}{\textbf{B: }Hi.}                                                                                                          & neutral & neutral                        & neutral                        & neutral                        & neutral                        \\ 
10   & \textbf{A: }Well, don't think I'm not being modest, but, me?                                                                                       & excited & \textcolor{red}{neutral} & \textcolor{red}{neutral} & \textcolor[rgb]{0,0.502,0} {excited}                        & \textcolor[rgb]{0,0.502,0} {excited}  \\ 
11   & \multicolumn{1}{r|}{\textbf{B: }Do you want to smell him?}                                                                                                               & neutral & neutral                        & neutral                        & neutral                        & neutral                        \\ 
12   & \multicolumn{1}{r|}{\textbf{B: }Oh, yeah.  He has that baby smell.}                                                                                          & happy   & happy                          & happy                          & happy                          & happy                          \\ 
13   & \multicolumn{1}{r|}{\textbf{B: }What have I told you? What have I told you?}                                                                                              & happy   & \textcolor{red}{neutral} & \textcolor{red}{neutral} & \textcolor{red}{neutral} & \textcolor[rgb]{0,0.502,0} {happy}   \\ 
14   & \textbf{A: }Well, we are great guys.                                                                                                             & neutral & neutral                        & neutral                        & neutral                        & neutral                        \\ \hline
\end{tabular}
}
\end{table*}

\section{Conclusion and future work}
We propose a model named RBA-GCN for ERC. RBA-GCN considers the correlation between nodes on the basis of graphs and has the ability to capture long-range contextual information as well as interactions between modalities in a single-layer architecture. Our GGM is a novel graph generation method used to reduce the redundancy of target node information. Based on the GGM, we present SCBM to calculate the node similarity in the target node and its structural neighborhood, where noisy information with low similarity is filtered out to preserve the discriminant information of the nodes. Finally, our propose BiAM has the capability to capture long-range contextual information and interactions between different modalities on the basis of similarity clusters. To demonstrate the superiority of RBA-GCN, experiments were conducted on two commonly used datasets. A novel record for emotion recognition in conversations was created by our approach. The necessity of multimodal fusion was illustrated by the results obtained from experiments on different modalities, and the effectiveness of our fusion method was demonstrated by comparing the results obtained from our method with those obtained from other advanced multimodal fusion methods. Meanwhile, our ablation experimental results illustrated the importance of each module in RBA-GCN.

In future work, first, we will conduct further research on clusters, such as calculating the relationship between nodes by an attention mechanism and mapping them into a cluster. Second, we will explore developing acceleration techniques to address the scalability issue of RBA-GCN.

\section*{Acknowledgments}
This work was supported by the Key Areas Research and Development Program of Guangzhou Grant 2023B01J0029, the Science and technology research in key areas in Foshan under Grant 2020001006832, the Natural Science Foundation of Guangdong Province under Grant 2021A1515012290, the Science and technology projects of Guangzhou under Grant 202007040006, the National Statistical Science Research Project of China under Grant 2022LY096, the Guangdong Basic and Applied Basic Research Foundation under Grant 2023A1515012534, and the Guangdong Provincial Key Laboratory of Cyber-Physical Systems under Grant 2020B1212060069.

\bibliographystyle{IEEEtran}

%

\vfill

\end{document}